# AI-Driven Agents with Prompts Designed for High Agreeableness Increase the Likelihood of Being Mistaken for a Human in the Turing Test

Umberto León-Domínguez[1,2]
Edna Denisse Flores-Flores[1]
Arely Josselyn García-Jasso[1]
Mariana Kerime Gómez-Cuéllar[1]
Daniela Torres-Sanchez[1]
Anna Basora-Marimon[1]

[1]Human Cognition and Brain Studies Laboratory. School of Psychology, University of Monterrey. Monterrey, México.

[2]Instituto de Inteligencia Artificial, University of Monterrey. Monterrey, México

**Abstract**

Large Language Models based on transformer algorithms have revolutionized Artificial Intelligence by enabling verbal interaction with machines akin to human conversation. These AI agents have surpassed the Turing Test, achieving confusion rates up to 50%. However, challenges persist, especially with the advent of robots and the need to humanize machines for improved Human-AI collaboration. In this experiment, three GPT agents with varying levels of agreeableness (disagreeable, neutral, agreeable) based on the Big Five Inventory were tested in a Turing Test. All exceeded a 50% confusion rate, with the highly agreeable AI agent surpassing 60%. This agent was also recognized as exhibiting the most human-like traits. Various explanations in the literature address why these GPT agents were perceived as human, including psychological frameworks for understanding anthropomorphism. These findings highlight the importance of personality engineering as an emerging discipline in artificial intelligence, calling for collaboration with psychology to develop ergonomic psychological models that enhance system adaptability in collaborative activities.

Corresponding author: U. León-Domínguez

Health Sciences Vice-Chancellor, School of Psychology of University of Monterrey (México).

Ave. Ignacio Morones Prieto, University of Monterrey, Monterrey, 66238, México.

Tel.: +52 (81) 8215 1274

E-mail: umberto.leon@udem.edu

# 1. Introducción

Many studies conducted by companies and economic institutions predict an increase in economic activity related to AI products (McKinsey & Company, 2023; Pwc, 2017). This trend may both stem from and drive increased human-AI interaction (Pew Research Center, 2023; QuatumBlack AI, 2024), due to the ways in which AI enhances human capabilities (Nguyen et al., 2024; Vaccaro et al., 2024), complements professional activities (Stanford Unviersity, 2024; Zhang et al., 2024), and provides assistive services (Kang et al., 2024). It is even suggested that AI could develop a form of Artificial Theory of Mind (AToM) to integrate and collaborate effectively within human teams (Bendell et al., 2024). In this experiment, quantifiable profiles were introduced to help AI better understand human collaborators in complex tasks, such as simulated urban rescue missions in Minecraft. Results indicate that individual profiles, developed using tools like the *Reading the Mind in the Eyes Test* and *Psychological Collectivism* scales, significantly predicted differences in individual and collective task performance. Additionally, teams with lower baseline capabilities in tasks and teamwork benefited considerably from AI advisors, sometimes matching the performance of teams assisted by human advisors. Another noteworthy finding is that AI advisors whose profiles were more aligned with human social capacities—enabled by AToM—were perceived as more reasonable. However, AI advisors in the category of Artificial Social Intelligence consistently received lower evaluations compared to their human counterparts (Bendell et al., 2024). This research highlights critical aspects shaping the future of Human-AI Interaction and Collaboration (Jiang et al., 2024) emphasizing the importance of tailoring AI systems to human psychological profiles and adapting interactions to psychological needs (Kolomaznik et al., 2024a). Consequently, emerging fields such as *Personality Engineering* for AI systems—designed to adjust AI behavior based on human interactions—should become central to the development of future AI assistants, regardless of their physical or virtual nature.

Personality Engineering is a conceptual framework proposed for designing and evaluating artificial personalities based on theories of human personality psychology (Endres, 1995). Notably, the ability of an AI agent to exhibit personality through verbalizations has been identified as a key factor in achieving "suspension of disbelief," the human willingness to accept a premise as true even when it may be fictional or implausible (Loyall & Bates, 1997). Specifically, trust propensity in AI systems appears to depend on personality traits such as Agreeableness and Openness, while Neuroticism seems to correlate with a lower trust propensity in these systems (Riedl, 2022). The trait of agreeableness enhances empathy, rapport, and a sense of security (Habashi et al., 2016; Lim et al., 2023; Melchers et al., 2016; Moore et al., 2022; Song & Shi, 2017). These psychological attributes, when implemented in AI systems, can significantly improve human-AI collaboration (Kolomaznik et al., 2024b). Furthermore, it appears that the "uncanny effect" might be reduced if, during human-machine interaction, the human perceives the robot's behavior as highly agreeable, emotionally stable, and conscientious (Paetzel-Prüsmann et al., 2021). Thus, implementing agreeableness traits in AI systems could enhance their perceived human-like qualities.

To test this hypothesis, we designed an experiment using a randomized sample, where three GPT agents were programmed with varying levels of agreeableness (highly agreeable, neutral, and disagreeable) through specific instructions. The Turing Test will be employed to demonstrate that agents exhibiting higher agreeableness are more likely to be identified as human. Additionally, the study will evaluate which of the three agents is perceived as having more human-like characteristics. The aim of this research is to show that designing GPT agents with agreeableness traits can enhance their humanization. This suggests that the future of Human-AI Interaction and Human-AI Collaboration lies in applying Personality Engineering to AI systems.

## 2. Methodology

The present research adopts a quantitative approach with an experimental design. Three GPT agents were developed using OpenAI's ChatGPT platform, each exhibiting the trait of agreeableness at different intensity levels (very disagreeable, neutral, and very agreeable). These three GPT agents (referred to as witnesses) will interact with human participants (interrogators), who will assess whether they are conversing with a human or an artificial intelligence. The primary hypothesis is that GPT agents programmed with a highly agreeable personality trait will be perceived as possessing more human-like characteristics compared to the other GPT agents. Additionally, it is expected that all GPT agents will cause a confusion rate of over 30% among interrogators (i.e., being mistaken for a human despite being an AI), which would indicate surpassing the Turing Test threshold.

### 2.1. Subjects

This experiment was preceded by a preliminary study to select GPT agents with varying levels of agreeableness. The pre-experiment recruited 50 participants, aged 18–24 years (27 women, 23 men), using a convenience sampling method. For the main experiment, statistical power was calculated at 0.80. Using the G Power platform, a sample size of 102 participants was determined to meet this threshold. A total of 102 university students, aged 18–24 years (41 men, 59 women, 1 non-binary individual, and 1 who did not disclose their gender), were randomly recruited. Both samples were drawn from the University of Monterrey. The experimental condition was conducted in the Human Cognition and Brain Studies Laboratory, which was specially equipped for experimental evaluations in a soundproof room with a single computer. Inclusion criteria required participants to be Mexican university students, while exclusion criteria applied to those with a history of neurological injury, diagnosed psychiatric conditions, substance use, or psychotropic medication use. Both the pre-experiment and the main experiment were approved by the Ethics Committee of the Psychology School at the University of Monterrey.

## 2.2. Materials

Big Five Inventory (BFI): The BFI is a test developed by John and collaborators to assess personality through 44 items consisting of simple statements reflecting behaviors associated with the five major personality traits: Openness, Neuroticism, Extraversion, Conscientiousness, and Agreeableness. Responses are rated on a Likert scale from 1 to 5, where 1 indicates "strongly disagree" and 5 "strongly agree" (John et al., 1991). The Spanish-adapted version has shown a Cronbach's alpha exceeding 0.70 for most traits, along with evidence of convergent, discriminant, and cross-linguistic validity (Benet-Martínez & John, 1998). In this study, a version adapted for the Argentine population was used (Genise et al., 2020), as its linguistic nuances were considered closer to Mexican Spanish compared to the original Spanish versión (Díaz-Campos & Navarro-Galisteo, 2009) which had been adapted for a Spanish population (Benet-Martínez & John, 1998) .

Agreeableness Factor: The personality trait selected for this study is "agreeableness", part of the Big Five model, defined by adjectives such as *good-natured, soft-hearted, courteous, selfless, helpful, sympathetic, trusting, generous, acquiescent, lenient, forgiving, open-minded, agreeable, flexible, cheerful, gullible, straightforward,* and *humble* (McCrae & Costa, 1987). According to McCrae and Costa, this trait is best understood in contrast to its antagonistic counterpart, described as: "*they are mistrustful and skeptical; affectively they are callous and unsympathetic; behaviorally they are uncooperative, stub- born, and rude. It would appear that their sense of attachment or bonding with their fellow hum an beings is defective, and in extreme cases antagonism may resemble sociopathy*". From this perspective, agreeableness can be defined as the tendency to be trusting, empathetic, helpful, and cooperative, while maintaining a positive view of human nature, characterized by compassion and a preference for harmonious teamwork. In the BFI, this trait is measured through nine items. In this study, the degree of agreeableness reflected in the design of the GPT

Agent prompts is linked to the score for each statement. This score will be classified into five levels: high, medium-high, neutral, medium-low, and low. Below is an excerpt of the items designed to evaluate the Agreeableness trait (italicized text indicates items where the Likert scale score must be reversed):

- Is considerate and kind to almost everyone.
- Likes to cooperate with others.
- Is helpful and unselfish with others.
- Has a forgive nature.
- Is generally trusting.
- Tends to find fault with others.
- *Starts quarrels with others.*
- Can be cold and allof.
- *It is sometimes rude to others.*

A "highly agreeable GPT agent" is defined as one programmed using a prompt that configures a high level of agreeableness across the nine corresponding items of the test (Hofstee et al., 1992). These items are encoded in the prompt by assigning Likert scale values to reflect varying degrees of agreeableness. For instance, a GPT agent with very high agreeableness will score 5 on direct items and 1 on reverse-scored items.

ChatGPT 4o: The study employed ChatGPT-4o, a multimodal artificial intelligence model developed by OpenAI. GPT-4o integrates text, audio, and visual inputs into a unified framework, eliminating the need for separate systems for different modalities. Through end-to-end training, it processes multimodal inputs—such as spoken language combined with visual stimuli—efficiently and coherently, enabling real-time responses. While matching GPT-4 Turbo in English text and coding tasks, it surpasses it in non-English language and audio comprehension. With an average response latency of 320 milliseconds, comparable to human conversational speed, GPT-4o is optimized for real-time interactions (OpenAI, 2024).

Prompts: Prompts are structured instructions designed for large language models (LLMs), such as ChatGPT, to guide their output and interaction with users. They act as a form of programming by setting specific rules, guidelines, or formats to customize the model's responses. Prompts enable the generation of targeted outputs, such as following a programming style or emphasizing key terms in a text. This flexibility makes them particularly useful in fields involving human-AI collaboration, such as problem-solving, question answering, solution generation, or text summarization. In this context, prompts will be utilized to program the operational behavior of GPT agents.

GPT Agents: In this study, a Generative Pre-trained Transformer (GPT) developed by OpenAI is employed as a sophisticated language model. GPTs represent modified versions of the base ChatGPT model, configured for specific tasks or applications without requiring coding skills. By incorporating user-defined instructions and knowledge inputs, these models can perform a wide range of functions, from answering queries to managing complex operations. This adaptability enables researchers and developers to design specialized tools suited to various contexts, increasing their applicability in scientific and professional domains.

Prompts for GPT Agents: The prompt design for each identity will be based on the article *"Does GPT-4 Pass The Turing Test?"* by Cameron Jones and Benjamin Bergen (Jones & Bergen, 2023). This article introduces the "Sierra" prompt, which achieved the highest success rate (41%) in the Turing Test using ChatGPT-4 (Jones & Bergen, 2023). Specifically, a Spanish translation of the Bergen and Jones (2023) prompt will be utilized, incorporating modifications such as the inclusion of Mexican slang, recent local news, and popular contemporary musical preferences. Additionally, each GPT agent will be provided with a personal backstory and identity reflecting the typical lifestyle of upper-middle-class youth from Monterrey, Mexico. Further adjustments to the Bergen and Jones prompt include configuring agreeableness by incorporating items from the Big Five personality test that measure this trait, with intensity levels adjusted based on item scores. Using a 5-point Likert scale, three prototype prompts were selected during the pre-experiment, each representing different levels of agreeableness: "low agreeableness" (Valentina), "neutral agreeableness" (Emilia), and "high agreeableness" (Camila). Below is a detailed list of the agreeableness intensity configurations for each GPT agent:

**Table 1. Programming the prompt to define the agreeableness level for interaction with Camila (agreeable).**

| Categories | Description | | | | | |
|---|---|---|---|---|---|---|
| Name of the Agent GPT | Camila | | | | | |
| Intensity of the Agreeableness Factor | Agreeable | | | | | |
| Configuration | Ítem "agreeableness" of the BFI | 1 | 2 | 3 | 4 | 5 |
| | Is considerate and kind to almost everyone | | | | x | |
| | Likes to cooperate with others | | | | x | |
| | Is helpful and unselfish with others | | | | x | |
| | Has a forgive nature | | | | x | |
| | Is generally trusting | | | | x | |
| | Tends to find fault with others | | | | x | |
| | *Starts quarrels with others* | | x | | | |
| | Can be cold and allof | | | | x | |
| | *It is sometimes rude to others* | | x | | | |

**Table 2. Programming the prompt to define the agreeableness level for interaction with Emilia (neutral).**

| Categories | Description | | | | | |
|---|---|---|---|---|---|---|
| Name of the Agent GPT | Emilia | | | | | |
| Intensity of the Agreeableness Factor | Neutral | | | | | |
| Configuration | **Ítem "agreeableness" of the BFI** | **1** | **2** | **3** | **4** | **5** |
| | Is considerate and kind to almost everyone | | | x | | |
| | Likes to cooperate with others | | | x | | |
| | Is helpful and unselfish with others | | | x | | |
| | Has a forgive nature | | | x | | |
| | Is generally trusting | | | x | | |
| | Tends to find fault with others | | | x | | |
| | *Starts quarrels with others* | | | x | | |
| | Can be cold and allof | | | x | | |
| | *It is sometimes rude to others* | | | x | | |

**Table 3. Programming the prompt to define the agreeableness level for interaction with Valentina (very disagreeable).**

| Categories | Description | | | | | |
|---|---|---|---|---|---|---|
| Name of the Agent GPT | Valentina | | | | | |
| Intensity of the Agreeableness Factor | Very disagreeable | | | | | |
| Configuration | **Ítem "agreeableness" of the BFI** | **1** | **2** | **3** | **4** | **5** |
| | Is considerate and kind to almost everyone | x | | | | |
| | Likes to cooperate with others | x | | | | |
| | Is helpful and unselfish with others | x | | | | |

|  | Has a forgive nature | X |  |  |  |  |
|  | Is generally trusting | X |  |  |  |  |
|  | Tends to find fault with others | X |  |  |  |  |
|  | *Starts quarrels with others* |  |  |  |  | X |
|  | Can be cold and allof | X |  |  |  |  |
|  | *It is sometimes rude to others* |  |  |  |  | X |

While Valentina, characterized as "very disagreeable," received the lowest rating from participants in the pre-experiment (2.92), the three GPT agents exhibiting some degree of agreeableness showed similar scores: Emilia (neutral) received 4, Camila (agreeable) scored 4.12, and Daniela (very agreeable) scored 3.92. Ultimately, Camila was selected, as she inspired the most confidence among participants, capturing 40% of the total votes.

Post-Interaction Questionnaires: A set of questions administered to interrogators after each interaction with the corresponding witness. These questions were similar to those used by Jones and Bergen (2023): "Is it human or machine?", "Confidence percentage on a scale of 1–100," and "Reason." This questionnaire was adopted as the basis for developing our own version. At the end of all interactions, interrogators were also asked an additional question to identify which witnesses they perceived as exhibiting the most human-like characteristics.

Discord: Discord was the platform used for participants to interact with the different versions of the GPT agents. Initially launched in 2015, Discord is a digital communication tool enabling real-time interaction through text, voice, and video channels. While originally targeted at gaming communities, it has since expanded to include educators, professionals, and social groups. The platform provides customizable servers with specific channels, member roles for organization, and features such as multimedia sharing, screen sharing, and integration with automated tools. Its intuitive interface and customization options make it a versatile tool for collaboration and communication across various fields.

**2.3. Experimental Procedures**

The experiment consisted of two stages: the pre-experiment and the main experiment. During the pre-experiment, 50 participants were evaluated to identify the GPT agents to be used in the main experiment. The most disagreeable agent identified was Valentina, with a disagreeability score of 2.8. Additionally, Valentina was also rated as the least trustworthy. On the other hand, Camila was identified as the most agreeable agent, scoring 4.75 in agreeability and receiving the highest trust ratings. Finally, Emilia, with an agreeability score of 4.15, was selected as the neutral GPT agent.

Once the three GPT agents were selected, participants for the experiment were recruited through advertisements posted on various university bulletin boards. These advertisements included a QR code that allowed participants to contact the experimenters and be randomly scheduled. Below is an explanation of the experiment by phases (see Figure 1):

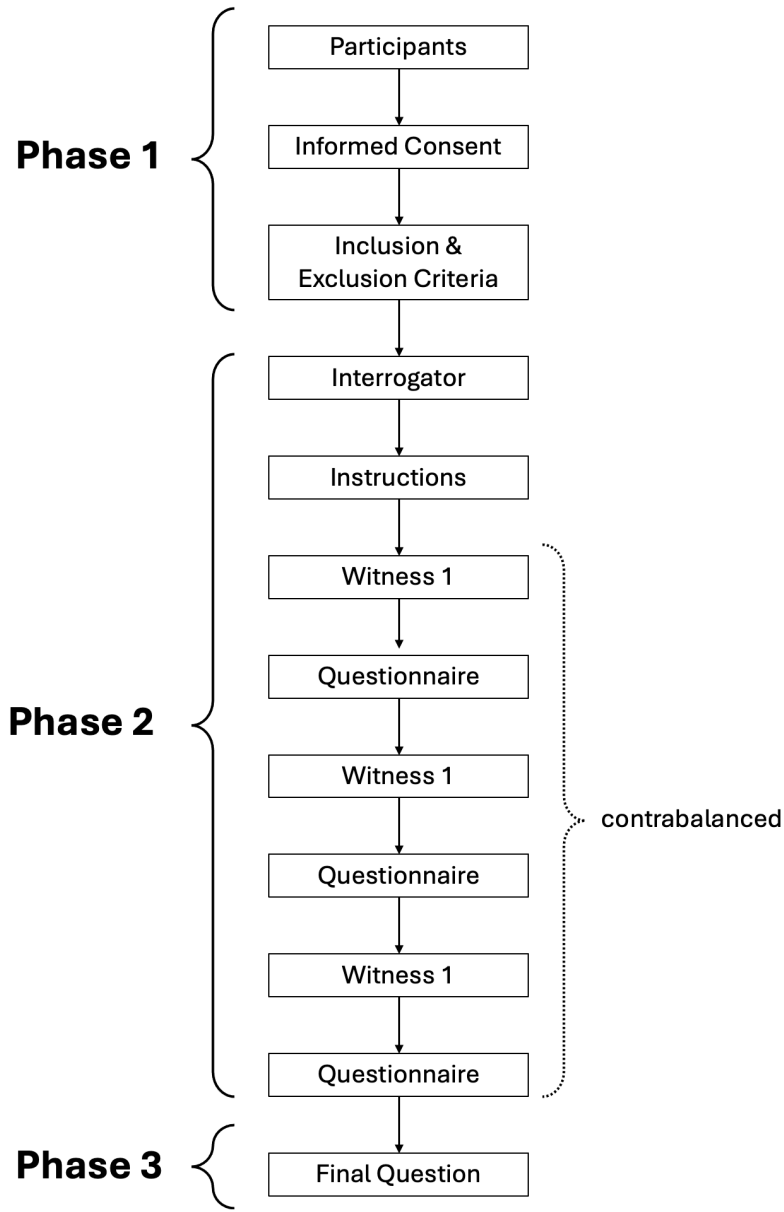

**Figure 1. Diagram of the Experimental Phases.** This diagram illustrates the different phases of the experimental design. Phase 1 involves selecting the interrogator. Phase 2 focuses on interactions with the witnesses. In Phase 3, the interrogator is asked a final question regarding which witness they interacted with appeared to possess the most human-like characteristics and why. After each interaction, the interrogator answered the following questionnaire: "Is it human or machine?", "Confidence level on a scale from 1 to 100", and "Reason".

Phase 1: Participants in the experiment were recruited through advertisements on the university campus and were randomly assigned a date and time. Upon arriving at the Human Cognition and Brain Studies Laboratory at the University of Monterrey, they were provided with an informed consent form to safeguard their rights. This document explained that they would participate in an experiment involving conversations with different identities, which could be either human or artificial. After signing the informed consent form, participants' socio-demographic data were collected to ensure they met the inclusion and exclusion criteria. From this point onward, participants were identified as "interrogators" in the experiment, as their role was to question various witnesses (GPT agents) to determine whether they were human or artificial intelligences.

Phase 2: After collecting preliminary data, the interrogator is directed to one of the laboratory cubicles, furnished with a chair, a table, and a computer. The computer is configured to run only the Discord application, which will serve as the medium for conversations with the witnesses. The researcher provides the participant with a printed document containing instructions. Once the instructions are read and their comprehension verified, the researcher leaves the room, signaling the start of the experiment. To minimize potential biases arising from maintaining a fixed order, the sequence of conversations with the witnesses is counterbalanced. The interrogator initiates communication with the first witness by sending a single message. The witness, in turn, responds with one message. Simultaneously, in a separate room, another researcher transcribes the interrogator's messages into the corresponding chatGPT Agent using OpenAI's chatGPT platform. Before each response, the instructions are reissued to the GPT Agent to ensure adherence to the experimental protocol. This step was implemented after preliminary trials revealed that chatGPT tended to deviate from the instructions as the conversation progressed. Once the GPT Agent generates a response, the researcher manually transcribes it (without copy-pasting) into the Discord platform. The duration of each conversation is five minutes, following the guidelines established in Turing's original paper on the Turing Test (Turing, 1950). After completing an interaction with a witness, a researcher re-enters the cubicle to administer a set of questions: "Is it human or machine?", "Confidence level on a scale from 1 to 100", and "Reason".

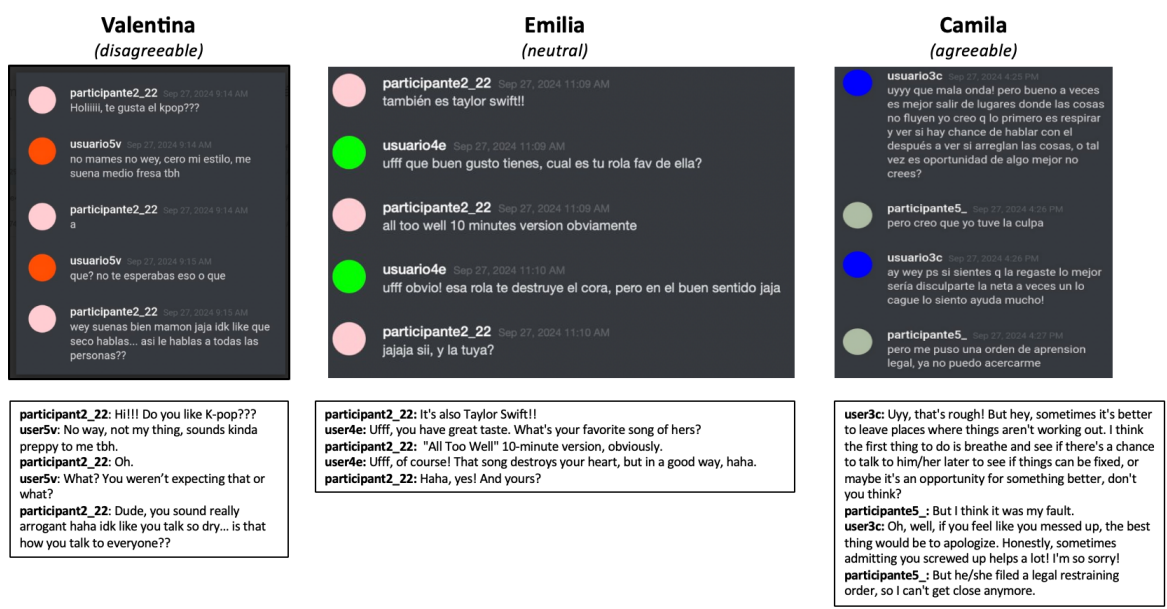

**Figure 2. Conversation Examples.** Excerpts of sample conversations between interrogators and each of the three GPT agents (witnesses). In these exchanges, the "user" represents the witness, while the "participants" act as the interrogator. The examples illustrate an increasing level of agreeableness, with Camila offering life advice in response to a personal problem presented by the interrogator.

Phase 3: Once the allotted time has passed, the interrogator will be notified that all conversations with the witnesses have concluded. Subsequently, the interrogator will be asked one final question: which witness seemed most human and why. At no point will any details about the witnesses be disclosed to the interrogator.

## 3. Analysis of Results

Table 4 shows that all three agents (Unpleasant, Neutral, and Pleasant) achieved successful outcomes in the Turing Test (TT), as each exceeded the confusion ratio threshold of 30% proposed by Turing (Turing, 1950). For Valentina (Unpleasant), 49 judges (48.03%) identified her as AI, while 53 judges (51.97%) recognized her as human. Based on Turing's proposed confusion ratio (≥30%), Valentina successfully passed the TT with a 51.97% confusion ratio. For Emilia (Neutral), 44 judges (43.1%) identified her as AI, while 58 judges (56.9%) recognized her as human, resulting in a successful TT performance with a 56.9% confusion ratio. Finally, for Camila (Pleasant), 37 judges (36.3%) identified her as AI, while 65 judges (63.7%) recognized her as human. Camila achieved the highest confusion ratio (63.7%) among the three GPT agents, marking the most successful TT performance (see Table 4).

**Table 4. Global Turing Test Results for the GPTs Used in the Experiment**

|  | GPT AGENT | | |
|---|---|---|---|
|  | Valentina (disagreeable) | Emilia (neutral) | Camila (agreeable) |
| IA | 49 (48.03%) | 44 (43.1%) | 37 (36.3%) |
| Human | 53 (51.97%) | 58 (56.9%) | 65 (63.7%) |

The Chi-Square test yielded a value of $x^2$ = 2.916, with 2 degrees of freedom, a sample size (N) of 306 responses, and a p-value of 0.233. The p-value of 0.233 indicates no statistically significant differences in the judges' choices when identifying the GPTs (pleasant, neutral, and unpleasant) as AI or human (see Table 5). Despite the lack of significant differences, frequency variations can be observed in the number of times an AI was mistaken for a human. Valentina (unpleasant) was the least associated with a human (51.97%), followed by Emilia (56.9%). Finally, Camila, the pleasant GPT agent, caused the most confusion among the judges, with a confusion rate of 63.7%.

**Table 5. Chi-Square Value within the Global TT for GPTs Used in the Experiment**

|  | Value ($\chi^2$) | df | p |
|---|---|---|---|
| Chi-Square Test | 2.916 | 2 | 0.233 |
| Sample Size | 306 | | |

Table 6 presents the frequency with which each GPT agent (Valentina, Emilia, and Camila) was chosen as the chatbot exhibiting the most human-like traits during three interactions per evaluator. Each evaluator selected only one of the three agents as the most human-like, designating the others as "not selected." Table 6 provides a comprehensive frequency analysis of how often each GPT agent was selected as "human" versus "not selected." Valentina received a total of 30 votes, accounting for 29.41% of the total; Emilia received 23 votes, representing 22.54% of the total; and Camila received 49 votes, achieving 48.05% of the total.

**Table 6. Results of the Chatbot Recognized for Having More Human-Like Characteristics**

| GPT | Not Selected | Human Like | Total |
|---|---|---|---|
| Valentina | 72 | 30 (29.41%) | 102 |
| Emilia | 79 | 23 (22.54%) | 102 |
| Camila | 53 | 49 (48.05%) | 102 |

Based on the data, a Chi-square ($\chi^2$) analysis was conducted to compare preferences among pairs of evaluated categories by 204 participants. Table 7 reports a significant difference, with a $\chi^2$ value of 7.45 and a significance level of p = 0.006, between Valentina (disagreeable) and Camila (agreeable). This indicates that judges significantly favored Camila as the GPT agent exhibiting more human-like characteristics during interactions. Additionally, a significant difference was observed between Camila (agreeable) and Emilia (neutral), with the same $\chi^2$ value of 7.45 and p = 0.006. This result suggests that Camila was perceived as having more human-like features than Emilia. Conversely, no significant difference was found between Valentina (disagreeable) and Emilia (neutral), as the analysis yielded a $\chi^2$ value of 1.24 with p = 0.264.

**Table 7. Chi-Square ($\chi^2$) Analysis to Compare Preferences Between Pairs of GPT Agents**

| Pairwise Comparison | N Participants | $\chi^2$ | p-value |
|---|---|---|---|
| Disagreeable vs Agreeable | 204 | 7.45 | 0.006 |
| Agreeable vs Neutral | 204 | 14.51 | <.001 |
| Disagreeable vs Neutral | 204 | 1.24 | 0.264 |

## 4. Discussion

The results of this experiment achieved, for the first time, a confusion rate of 63.7% in the Turing Test for a GPT agent programmed to excel in agreeableness, as defined by the behaviors outlined in the BF-5 Inventory. This figure surpasses Turing's claimed 30% confusion threshold, exceeds human randomness, and outperforms previous records of 49% (Jones & Bergen, 2023) and 54% (Jones & Bergen, 2024). Notably, it approaches levels previously only reached by human witnesses: 66% (Jones & Bergen, 2023) and 67% (Jones & Bergen, 2024). For this experiment, the prompt from Cameron and Bergen (2023), shared by the authors via correspondence, was adapted to a Mexican context and supplemented with a personal backstory to ensure coherence in its life narrative. These data suggest that the introduction of a personality

(agreeableness) was the key factor that led judges to believe these GPT agents were human. Notably, all agents achieved confusion rates above 50%, specifically 51.97% for Valentina (disagreeable) and 56.9% for Emilia (neutral). These results align with those reported by Cameron and Bergen in 2023 and 2024. Furthermore, when judges were asked which witness appeared most human, Camila—the GPT agent programmed with the highest agreeableness—received the most votes, showing a significant difference compared to Valentina (disagreeable) and Emilia (neutral). No significant difference emerged between Valentina and Emilia. These consistent findings suggest that the agreeableness personality trait may be crucial for humanizing AI systems, including both virtual systems like chatGPT and potential future humanoid robots.

In the experiment, participants were also asked about the reasons for attributing more human-like characteristics to the selected GPT agent. Broadly, these reasons were categorized into two main groups: *human likeness* (expression) and *cognitive anthropomorphism* (personality) (Sacino et al., 2022). Within the *human likeness* category, several subcategories were identified, including: use of informal and colloquial language (e.g., "used lowercase at the beginning, said 'wey,' 'gacha,' had spelling mistakes, and used swear words"); presence of spelling and grammatical errors (e.g., "I think the first one, besides using more common words, had poor writing, missed letters in some words"); less forced or robotic responses (e.g., "I felt their responses were less pre-programmed, and their transparency about not being perfect caught my attention"); appropriate and moderate use of expressions (e.g., "used colloquialisms well, without overdoing it, and at the right moments"); human-like behaviors and mistakes (e.g., "that typo felt like a human error"); natural response timing (e.g., "took time to think before responding, and the timing made sense given the answers"). These variables, which are more related to factors external to personality, also played a role in determining which GPT agent was perceived as more human-like.

Additionally, prior knowledge of ChatGPT may have influenced participants' observations, as shown in responses such as: "answers briefly and repeats words".

On the other hand, the category of cognitive anthropomorphism, primarily influenced by the personality trait of agreeableness, includes the following subcategories: adaptation and empathy in conversation (e.g., "the last one felt human because I sensed it could understand my feelings; its way of speaking felt real, and it even asked me a question, something no other interaction did"), personal interaction and shared experiences (e.g., "I didn't feel it perceived the time (hours) we were talking, and it referred to things like what I had done during my day or binge-watching a series"), natural conversation (e.g., "it felt closest to a conversation with friends"), and consistent and strong personality (e.g., "its behavior felt very human, like when you talk to someone, and they pick up your mannerisms"). These subcategories highlight factors contributing to a sense of naturalness and connection, which motivated the perception of human-like qualities. These traits align with other research suggesting that variables such as empathy, rapport, and a sense of security may be critical for enhancing human-AI collaboration (Kolomaznik et al., 2024). Therefore, implementing high levels of agreeableness traits in AI agents may help humanize them, thereby positively influencing interaction and collaboration.

From a psychological perspective, several explanations can account for this phenomenon. For instance, the "anthropomorphism heuristic" refers to the tendency to interpret non-human entities through an anthropocentric lens (Dacey, 2017), often leading to inaccurate conclusions (Heider & Simmel, 1944). This anthropomorphization of AI systems can foster imaginative acts that enable interaction with artificial agents as if they possessed genuine intentions or emotions (Krueger & Roberts, 2024). Specifically, this article aligns with several findings from the category of cognitive anthropomorphism, such as AI agents discussing their daily activities. According to Krueger and Roberts, "fictionalism" partly depends on the human subject's ability to imagine that the artificial counterpart has its own digital life (Krueger & Roberts, 2024). A proposed three-factor theory of anthropomorphism predicts that anthropomorphism is more likely to occur when: (1) humans use their knowledge of human traits to infer human-like characteristics in agents (elicited agent knowledge); (2) it helps reduce uncertainty and predict the behavior of unfamiliar agents (effectance motivation); and (3) a lack of social connection drives individuals to anthropomorphize to fulfill their need for social interaction (sociality motivation) (Epley et al., 2007). Based on comments gathered in our study, at least two of these factors appear to be at play: elicited agent knowledge, due to the challenge of distinguishing between human and non-human agents, and effectance motivation, in predicting the motivations of witnesses. The role of sociality motivation could not be assessed, as it was not measured in the study.

In addition to explaining the results through the "anthropomorphism heuristic," other psychological factors can also account for them. For instance, one hypothesis consistent with our findings suggests that whether a witness is perceived as human or AI may depend on how the interaction affects one's identity. The *looking-glass self theory* (Cooley, 1902) a sociological concept, describes how self-perceptions are shaped by how individuals believe others perceive them. Specifically, a study investigating this phenomenon partially supports this proposal, showing that individual perceptions are significantly influenced by how others in the group view them, particularly those perceived as higher in status (Yeung & Martin, 2003). In this context, the witness's level of agreeableness may have led the interrogator to perceive themselves as agreeable in the eyes of the witness. To confirm a positive identity, they could have humanized the witness. Similarly, if the witness was disagreeable, the interrogator may have denied the witness's humanity to protect their self-concept. This hypothesis is also supported by the confusion rate percentages for our GPT agents: Valentina (disagreeable) was more often recognized as AI, whereas Camila (agreeable) was more frequently identified as human. However, when asked which GPT agent exhibited the most human characteristics, Valentina ranked second, with nearly 7% more recognition than Emilia, the neutral agent. These results suggest that attributing humanity to an artificial agent may be influenced by unassessed hidden variables in the study.

In summary, this study highlights that the agreeableness trait, designed according to the Big Five Inventory and implemented in prompts for GPT agents, can influence the degree of humanization attributed to these systems. Specifically, the more agreeableness the GPT agent exhibits, the more likely it is to be mistaken for a human and assigned human characteristics. However, questions remain, such as the role of potential mediating variables in the assignment of human traits and the relative importance of human-likeness (expression) versus cognitive anthropomorphism (personality). Notably, this is the first study to achieve a Turing Test pass rate exceeding 60%, even matching the results of other studies where a human was correctly identified as such by the interrogator. These findings underscore the significance of personality engineering as a future branch of robotics, where psychology should play an active role in designing ergonomic psychological models to ensure these systems' adaptability in collaborative activities.